\documentclass{article}

\usepackage[preprint,nonatbib]{neurips_2019} 

\usepackage[utf8]{inputenc} 
\usepackage[T1]{fontenc}    
\usepackage{hyperref}       
\usepackage{url}            
\usepackage{booktabs}       
\usepackage{amsfonts}       
\usepackage{nicefrac}       
\usepackage{microtype}      

\usepackage[square,numbers]{natbib}
\usepackage{amsmath}
\usepackage{mathtools}
\usepackage{xcolor}
\usepackage{tabularx}
\usepackage{booktabs}
\usepackage{multirow}
\usepackage{enumitem}
\usepackage{subcaption}

\DeclarePairedDelimiter\ceil{\lceil}{\rceil}
\DeclarePairedDelimiter\floor{\lfloor}{\rfloor}

\def\placeholder#1#2{\fbox{\vbox to #2{\vss\hbox to #1{\hss~}\vss}}}

\newcommand{\header}[1]  {\par\penalty100\vspace*{0pt}\noindent{\bf #1}\hspace{1em}}

\title{Accelerating Large-Kernel Convolution Using Summed-Area Tables}

\author{%
    {\bf Linguang Zhang \qquad Maciej Halber \qquad Szymon Rusinkiewicz}\\
    Department of Computer Science\\
    Princeton University\\
}

\begin{document}

\maketitle

\begin{abstract}
Expanding the receptive field to capture large-scale context is key to obtaining good performance in dense prediction tasks, such as human pose estimation. While many state-of-the-art fully-convolutional architectures enlarge the receptive field by reducing resolution using strided convolution or pooling layers, the most straightforward strategy is adopting large filters. This, however, is costly because of the quadratic increase in the number of parameters and multiply-add operations. In this work, we explore using learnable \emph{box filters} to allow for convolution with arbitrarily large kernel size, while keeping the number of parameters per filter constant. In addition, we use precomputed \emph{summed-area tables} to make the computational cost of convolution independent of the filter size. We adapt and incorporate the box filter as a differentiable module in a fully-convolutional neural network, and demonstrate its competitive performance on popular benchmarks for the task of human pose estimation.
\end{abstract}

\section{Introduction}

Fully-convolutional neural networks have seen success in numerous dense prediction tasks since \citet{long2015fully} adapted architectures that were originally used for image classification. In many cases, achieving high performance requires a large receptive field, and recent network architectures such as the popular ResNet family~\cite{he2016deep} use a large number of $3\times3$ convolution layers coupled with downsampling to capture large-scale contextual information. While this strategy does increase the receptive field, the use of downsampling prevents the network from generating high-resolution output.  Omitting downsampling while still only relying on $3\times3$ convolution is generally infeasible, since the network would need to be significantly deeper to achieve the same receptive field.

Previous work has proposed many ways to deal with this inherent conflict between increasing the receptive field and providing high-resolution output. An immediate solution is to add deconvolution (transposed convolution) layers to upsample the results. \citet{xiao2018simple} demonstrate that appending deconvolution layers to a ResNet backbone leads to a straightforward solution for restoring the resolution. Similarly, encoder-decoder and U-shaped networks, which are widely used for dense prediction tasks~\cite{newell2016stacked,ronneberger2015u}, also rely on deconvolution to produce high-resolution predictions. 

Because deconvolution adds significant complexity, an alternative is simply to expand the filter size.  Naive implementation of this strategy, however, leads to quadratic growth in both the number of operations and the number of parameters, leading to slower computation and greater susceptibility to over-fitting.
Dilated convolution~\cite{yu2015multi,chen2018deeplab} uses filters that effectively have a larger size, but utilize zero-padding to get by with fewer operations and parameters.  The drawback of dilated convolution, however, is the presence of ``gridding'' artifacts that can degrade performance in some applications.

\begin{figure}[t]
    \centering
    \includegraphics[width=\textwidth]{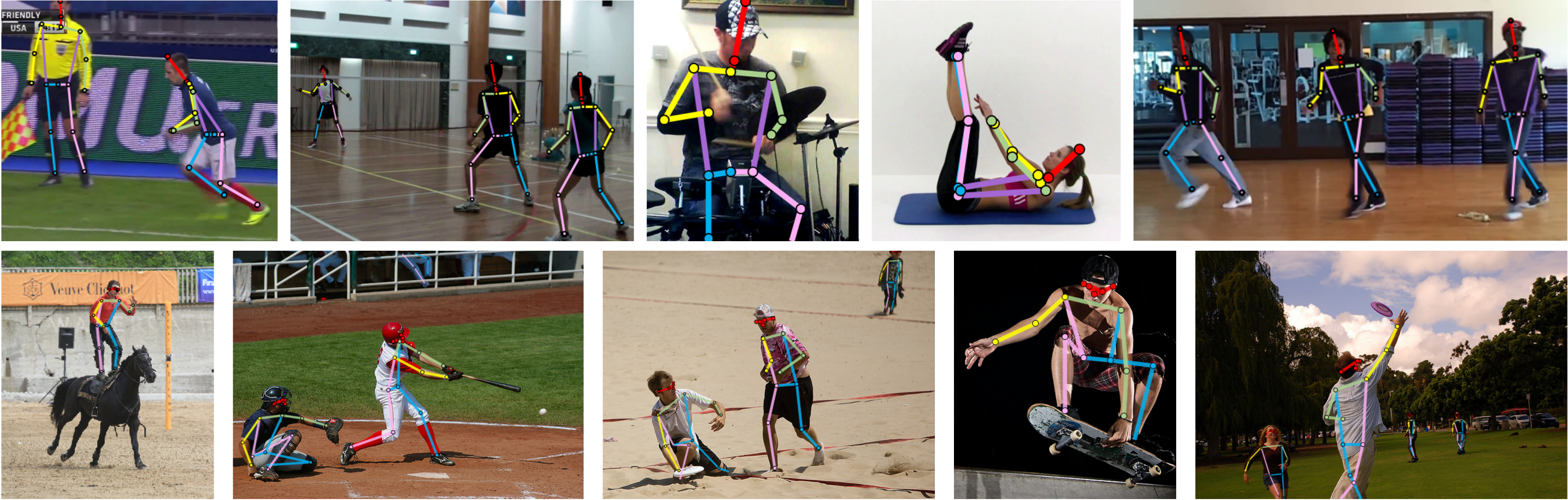}
    \caption{Qualitative results for a dense prediction task\,---\,human pose estimation\,---\,implemented using the proposed method. 
    \textbf{Top:}
    Results on the test set of MPII Human Pose.
    \textbf{Bottom:}
    Results on the \texttt{test-dev2017} set of Microsoft COCO.
    The proposed method is able to deal with a variety of cases, including multiple people at different scales, self occlusions, and non-standard poses.}
    \label{fig:qualitative_eval}
\end{figure}

In this paper, we explore large-kernel convolution using a classical approach that nevertheless has had limited application in the context of neural networks.  Specifically, we exploit \emph{Summed-Area Tables} (SATs), also known as \emph{integral images}, which enable the integral of an arbitrarily-sized rectangular region to be computed in constant time~\cite{lewis1995fast}.
%
SATs are therefore ideal for convolving an image with a box filter, since each output pixel requires a constant number of operations. Also, because a box filter can be parameterized using only four variables\,---\,the $(x,y)$ location and size of the box in the kernel\,---\,the number of parameters for a box filter is independent of the kernel size. The resulting efficiency, in terms of both computational cost and number of parameters, makes box filters well-suited for tasks that require both a large receptive field and high-resolution output (see Figure~\ref{fig:qualitative_eval}).

One possible drawback of box filters is the difficulty of detecting complex spatial patterns. In the context of neural networks, this issue is alleviated, as the intermediate feature maps often contain many channels. Almost any complex kernel can be well approximated by combining a large and diverse collection of box filters with appropriate weights. This observation implies that one can first convolve the image with a collection of box filters, then linearly combine the filtered images with different weights, which approximates convolving the image with various complex kernels. We show how to achieve this with depth-wise convolution followed by 1$\times$1 convolution.

Our main contribution is a lightweight, fully-convolutional network that uses SATs and box filters to perform large-kernel convolution, efficiently combining high-resolution output with wide receptive fields for pixel-level prediction tasks.
To enable this, we show how to implement box convolution in an end-to-end differentiable setting, in which the gradient of a box filter with respect to its parameters (position and size) must be continuous.  The resulting formulation uses sub-pixel positioning for all four corners of the box, and is computationally equivalent to conventional $4 \times 4$ convolution.

The specific task on which we demonstrate our results is human pose estimation, which can be posed as a high-level keypoint detection problem~\cite{lin2014microsoft}. 
This choice is motivated by the effectiveness of SATs in tasks such as local feature detection~\cite{bay2006surf} and face detection~\cite{viola2004robust}.
We show that the proposed architecture can achieve competitive results on popular benchmarks~\cite{andriluka20142d,lin2014microsoft} with only 1.85M parameters (an order of magnitude less than previous methods) and at a lower computational cost. We also demonstrate that our network generalizes better when the training data is limited.
To summarize, our contribution is three-fold:
\begin{itemize}[noitemsep,topsep=0pt]
  \item Deriving a generalizable solution to obtaining gradients for simple kernels that leverage summed-area tables for acceleration.
  \item Designing a lightweight fully-convolutional network that produces pixel-level prediction with a large receptive field.
  \item Demonstrating competitive performance on human pose estimation, with improved generalization when trained on limited data.
\end{itemize}

\section{Related Work}
\paragraph{Strategies to Enlarge Receptive Field} 
For visual recognition systems based on convolutional neural networks, having a sufficiently large receptive field to aggregate spatially distant information is critical.
For example, image classification networks~\cite{deng2009imagenet,krizhevsky2012imagenet,he2016deep} typically rely on downsampling (e.g., pooling and strided convolution) interleaved with stacks of small kernels,
exploiting the fact that downsampling essentially ``amplifies'' the receptive field of \emph{every} following kernel.


This downsampling, while generally acceptable for tasks such as classification that produce sparse or ``global'' output, is inappropriate for dense prediction tasks such as semantic segmentation or heatmap-based human pose estimation. To produce pixel-level predictions, a straightforward strategy is to couple downsampling with upsampling (e.g., deconvolution).
U-net and stacked-hourglass architectures~\cite{ronneberger2015u,newell2016stacked} adopt such a downsampling-upsampling strategy, together with skip connections to retrieve high-frequency details from early stages.
Cascaded Pyramid Networks~\cite{chen2018cascaded} obtain pixel-level predictions by fusing feature maps of different resolutions, which are produced by a progressive downsampling backbone network (e.g., ResNet~\cite{he2016deep}).

Avoiding downsampling altogether, while maintaining a large receptive field and computational efficiency, can be accomplished using an approach such as dilated convolution~\cite{yu2015multi,chen2018deeplab}. This technique inserts zeros between kernel elements, which produces the benefits of large-kernel convolution while unfortunately also resulting in gridding artifacts.
The differentiable box filter used in this work is another instance of large-kernel convolution. Compared to dilated convolution, which only sparsely utilizes pixels within the kernel, a box filter can leverage more pixels and produce a smoother result, depending on the size of the learned box.

\paragraph{Summed-Area Tables in Computer Vision}
Accelerating convolution with simple kernels using summed-area tables has a long history in computer vision. 
A classic application that popularized SATs is the efficient face detection method by~\citet{viola2004robust}.
Blob detection can also leverage summed-area tables for efficient image filtering when the Laplacian of Gaussian (LoG) is approximated with box filters~\cite{bay2006surf}.
\citet{trzcinski2012efficient} showed that complex, large kernels can be approximated with a few box filters, speeding up linear projection.

These ideas naturally lead to the concept of SAT-accelerated box filters as first-class elements in deep convolutional networks, and this was first explored in the work of \citet{burkovbox}.  The authors show that box filters can replace dilated convolution, leading to improved performance in the context of two existing lightweight networks for semantic segmentation~\cite{paszke2016enet,romera2017erfnet}.  However, that work has several drawbacks that are not shared by our (independently-developed) implementation.
First and most importantly, the implementation of \citeauthor{burkovbox} leads to discontinuities with respect to filter size and position, caused by a failure to perform sub-pixel sampling from the SAT at box corners.  In contrast, we perform correct sub-pixel sampling in both the forward and back-propagation passes, and demonstrate that this strategy easily generalizes to kernels other than single boxes. 
Second, the size of each box in their implementation can be larger than the image itself, and regularization is used to encourage the boxes to shrink.  We avoid regularization of box size, which can bias the training process, and instead impose a maximum-size constraint that may be considered analogous to the kernel size in standard convolution.
Third, they observe that the learned boxes are unintuitively symmetric with respect to the vertical axis, and that this is not a consequence of data augmentation. We have not observed the same phenomenon, suggesting that our correct sub-pixel sampling in both the forward pass and gradient estimation introduces less bias.

\section{Approach}
\subsection{Fast Convolution with Summed-Area Tables}
\label{sec:SAT_conv}

The summed-area table can be used to accelerate image convolution with kernels
that only involve rectangular summations. For simplicity, we begin by considering
how to efficiently convolve an image (a single-channel feature map) with the
simplest box filter. We imagine a filter kernel of maximum size $k$ that is
zero everywhere except for a rectangular sub-region
(i.e., box) filled with ones. The
extent of the box is specified by four integers $x_l, x_h, y_l, y_h
\in [0, k)$, and we define the filter $g$ as:
\[
    g_{i, j} = 
    \begin{cases} 
      1 & x_l \leq i \leq x_h ~~\text{and}~~ y_l \leq j \leq y_h\\
      0 & \text{otherwise.}
   \end{cases}
\]
The above filter can be used to perform general convolution. Denoting the input
image as $\mathcal{I}$ and the output image as $\mathcal{O}$, each pixel in the
output image is computed as:
\begin{equation}
    \mathcal{O}_{x,y} 
        = \sum_{i=0}^{k-1}\sum_{j=0}^{k-1}\mathcal{I}_{x+i, y+j} \, g_{i, j}
        = \sum_{i=y_l}^{y_h}\sum_{j=x_l}^{x_h}\mathcal{I}_{x+i, y+j} .
\end{equation}
The above equation costs at most $k^2$ \texttt{multadd} (multiply-add) 
operations to
compute each output pixel. However, since every output pixel is the
sum of a rectangular region, a precomputed summed-area table can be used to
achieve constant-time summation at each output location.
The value at location $(x, y)$ in the summed-area table $\mathcal{S}$ equals
the sum of all pixels above and to the left of $(x, y)$ in the input image
(including $\mathcal{S}_{x,y}$ itself). Therefore the sum of all pixels enclosed
in the box $[x + x_l, x + x_h]\times[y + y_l, y + y_h]$ can be efficiently
computed by sampling at the four corners of the box in the summed-area table:
\begin{equation}
    \mathcal{O}_{x, y} = \mathcal{S}_{x + x_h + 1, y + y_h + 1} +
                         \mathcal{S}_{x + x_l, y + y_l} -
                         \mathcal{S}_{x + x_l, y + y_h + 1} -
                         \mathcal{S}_{x + x_h + 1, y + y_l} .
\end{equation}
The summed-area table can be efficiently computed in a single pass over the
input image on the CPU, or using row/column parallelization on a GPU. Note that the
precomputation cost, relative to the cost of convolution, is quickly amortized when
the box becomes larger.

We have so far assumed that the box is aligned with the input, which is
a discrete lattice. In a neural network, we would like to make
$x_l, x_h, y_l, y_h$ learnable parameters instead of manually chosen integers,
and the resulting continuous optimization naturally leads to non-integer
coordinates. One could simply round the sampling points to the nearest integer-valued
coordinates, but the rounding operator is unfortunately not differentiable. 
\citet{burkovbox} update parameters using approximate gradients derived
through other means, such as normalizing the sum by the area of the
box~\cite{burkovbox}, but they do not directly address the discontinuity of
sampling.  We instead \emph{interpolate} among the four nearest values in the SAT
to accommodate non-integer coordinates.  While any differentiable interpolation function
could be used, we adopt bilinear interpolation in our implementation.

\begin{figure}
    \centering
    \newcolumntype{C}{>{\centering\arraybackslash}X}
    \setlength\tabcolsep{30pt}
    \begin{tabularx}{\hsize}{CC}
        \includegraphics[width=\hsize]{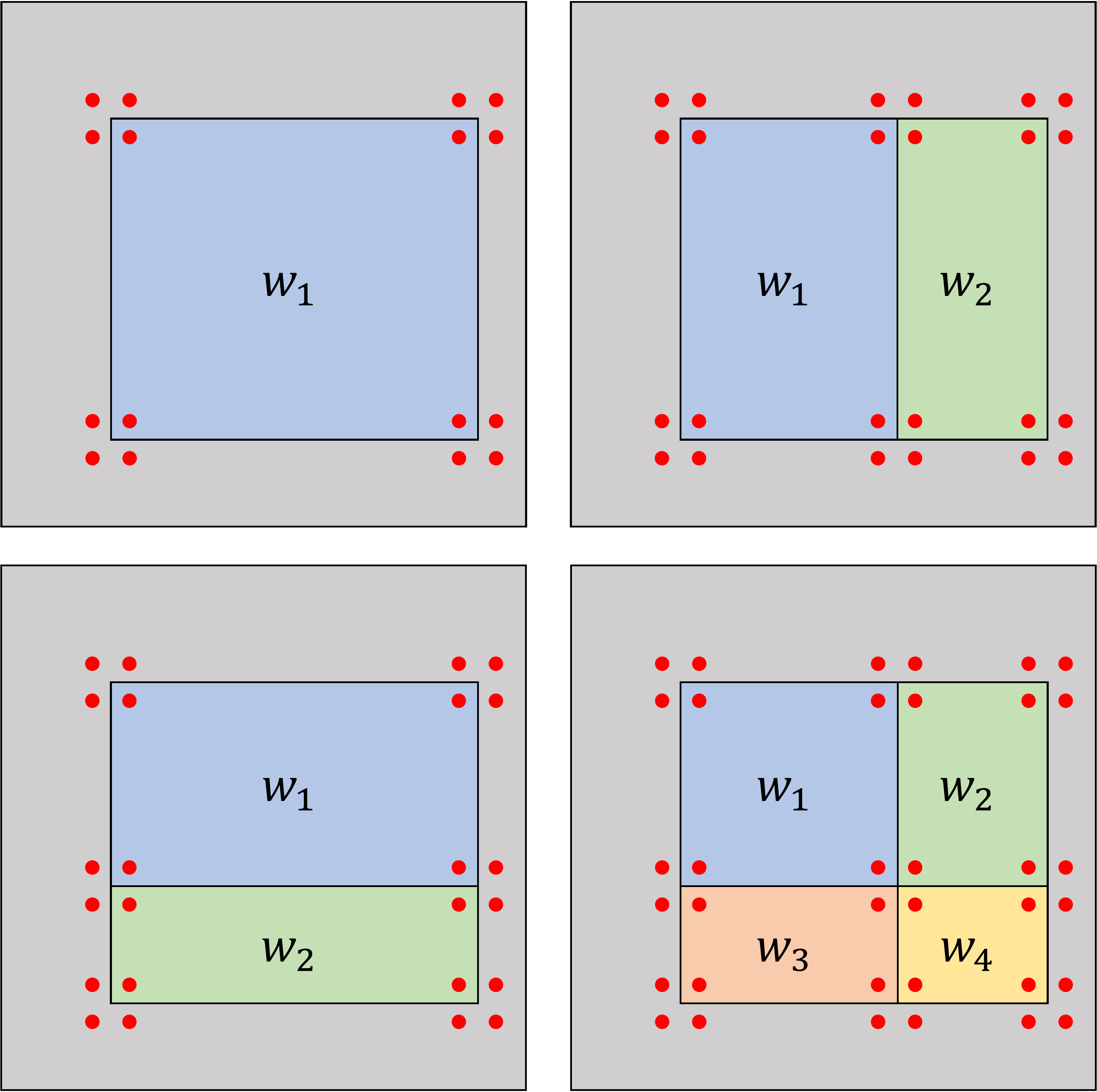}&
        \includegraphics[width=\hsize]{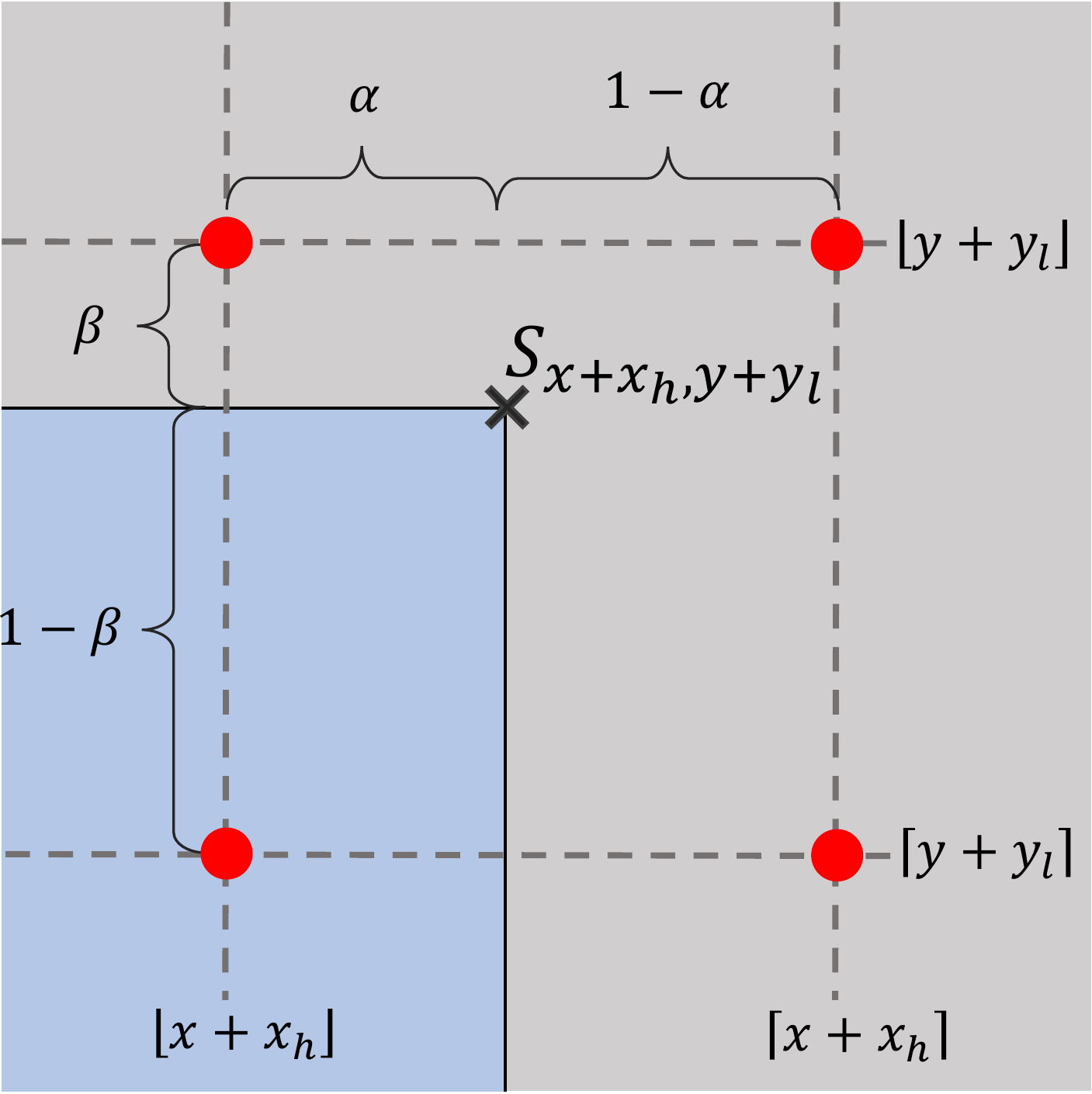}
    \end{tabularx}
    \caption{\textbf{Left:} a simple box filter, together with variants obtained through kernel splitting.
    Red dots indicate locations at which the SAT is sampled.
    \textbf{Right:} bilinear interpolation is performed at each corner, with the weights $\alpha$ and $\beta$
    remaining constant over the course of a single convolution.}
  \label{fig:box_conv_explanation}
\end{figure}

While this use of interpolation leads to a greater number of accesses to the SAT
to compute the convolution, we note that the relative cost could be
reduced by using more complex kernels.  For example, consider the four variants
illustrated in Figure~\ref{fig:box_conv_explanation}, left.  The use of ``kernel
splitting'' allows the use of kernels in which the box is divided into, say, 2 or 4
pieces with different weights.
Sampling from the SAT (illustrated by red dots) along an edge shared by two split
boxes can be performed only once, saving parameters and computation.

Regardless of whether a single box or a split kernel is used, we need to
compute the gradient of each convolved pixel with respect to box parameters
and sampled pixels.
Inspired by the Spatial Transformer~\cite{jaderberg2015spatial}, we use a
differentiable interpolation function to obtain the (sub-)gradients when
sampling from the SAT.  The
value sampled from the SAT at (possibly) non-integer-valued coordinates
$(\hat{x}, \hat{y})$ is computed as (see Figure~\ref{fig:box_conv_explanation}, right):
\begin{equation}
    \mathcal{S}_{\hat{x}, \hat{y}} = 
    (1-\alpha)(1-\beta) \cdot \mathcal{S}_{\floor{\hat{x}}, \floor{\hat{y}}} + 
    \alpha (1-\beta)    \cdot \mathcal{S}_{\ceil{\hat{x}}, \floor{\hat{y}}} +
    (1-\alpha)\beta     \cdot \mathcal{S}_{\floor{\hat{x}}, \ceil{\hat{y}}} +
    \alpha \beta        \cdot \mathcal{S}_{\ceil{\hat{x}}, \ceil{\hat{y}}},
\end{equation}
where $\floor{.}$ and $\ceil{.}$ are the floor and ceiling operators, 
$\alpha = \hat{x} - \floor{\hat{x}}$, and $\beta = \hat{y} - \floor{\hat{y}}$.
The above equation is continuous and differentiable in the interpolation
neighborhood, and the partial derivative of $\mathcal{S}_{\hat{x}, \hat{y}}$
with respect to $\hat{x}$ can be written as:
\begin{equation}
    \frac{\partial \mathcal{S}_{\hat{x}, \hat{y}}}{\partial \hat{x}} = 
        -(1-\beta) \cdot \mathcal{S}_{\floor{\hat{x}}, \floor{\hat{y}}} + 
        (1-\beta)  \cdot \mathcal{S}_{\ceil{\hat{x}}, \floor{\hat{y}}}
        -\beta     \cdot \mathcal{S}_{\floor{\hat{x}}, \ceil{\hat{y}}} +
        \beta      \cdot \mathcal{S}_{\ceil{\hat{x}}, \ceil{\hat{y}}}.
\end{equation}
The partial derivative with respect to $\hat{y}$ can be computed similarly.
The partial derivative of $\mathcal{S}_{\hat{x}, \hat{y}}$ with respect to each
sampled pixel, for example, $\frac{\partial \mathcal{S}_{\hat{x},
\hat{y}}}{\partial\mathcal{S}_{\floor{\hat{x}}, \floor{\hat{y}}}}$, can be
trivially computed as $(1-\alpha)(1-\beta)$.

An important observation is that shifting $\hat{x}$ or $\hat{y}$ by an integer
number of pixels does not change the value of $\alpha$ or $\beta$. For instance,
$\alpha = \hat{x} - \floor{\hat{x}} = \hat{x} + 1 - \floor{\hat{x} + 1}$.
This implies that we can precompute the weights for bilinear interpolation, which
involve only $\alpha$ and $\beta$, if we assume the most common scenario that
the convolution is performed with an integer-valued stride (e.g., stride = 1). 
With precomputed interpolation weights, sampling a pixel from the SAT
costs 4 \texttt{multadds}. Therefore, the total number of \texttt{multadds} per pixel spent
on computing the region sum becomes 16 when adopting bilinear interpolation.
In other words, since we sample 16 pixels and multiply with 16 precomputed
weights, our differentiable box filter is \emph{computationally equivalent} to a
$4\times4$ conventional convolution.
The weights can also be precomputed when kernel splitting is applied: the weight
of each split box can be combined with its interpolation weights.

\subsection{Implementation Details}
While a single box filter is simple, multiple box filters can be linearly
combined to approximate more complex kernels~\cite{trzcinski2012efficient}. To
leverage this property, we incorporate the differentiable box filter into a
depth-wise convolution layer in which each box filter is convolved with one channel of
the input feature map.
In many deep learning frameworks (e.g.,
Caffe~\cite{jia2014caffe} or PyTorch~\cite{paszke2017automatic}), standard
(non depth-wise) convolution is usually converted to matrix multiplication,
which allows the use of well-optimized General Matrix Multiply
(\texttt{gemm}) implementations. However, we have observed that
implementing depth-wise convolution in this manner is inefficient, because
laying out the patches as a matrix and invoking \texttt{gemm} introduces memory and
computational overhead. We instead parallelize the
computation of each output pixel directly, which is significantly faster in practice. 
Our CUDA implementation, wrapped as a standalone PyTorch layer, is included in the supplemental material.

We re-parameterize the four box coordinates $x_l, x_h, y_l, y_h$ into the $[-1, 1]$
range, relative to a maximum box size, and convert them back
during forward propagation. We clip
the parameters if they grow beyond $[-1, 1]$, and ensure that $x_l \leq x_h$ and $y_l \leq x_h$
after each iteration.  The parameters are initialized uniformly in the range
$[-0.5, 0.5]$ to prevent frequent clipping at the beginning of training.

\begin{figure}[t]
  \centering
  \includegraphics[width=\hsize]{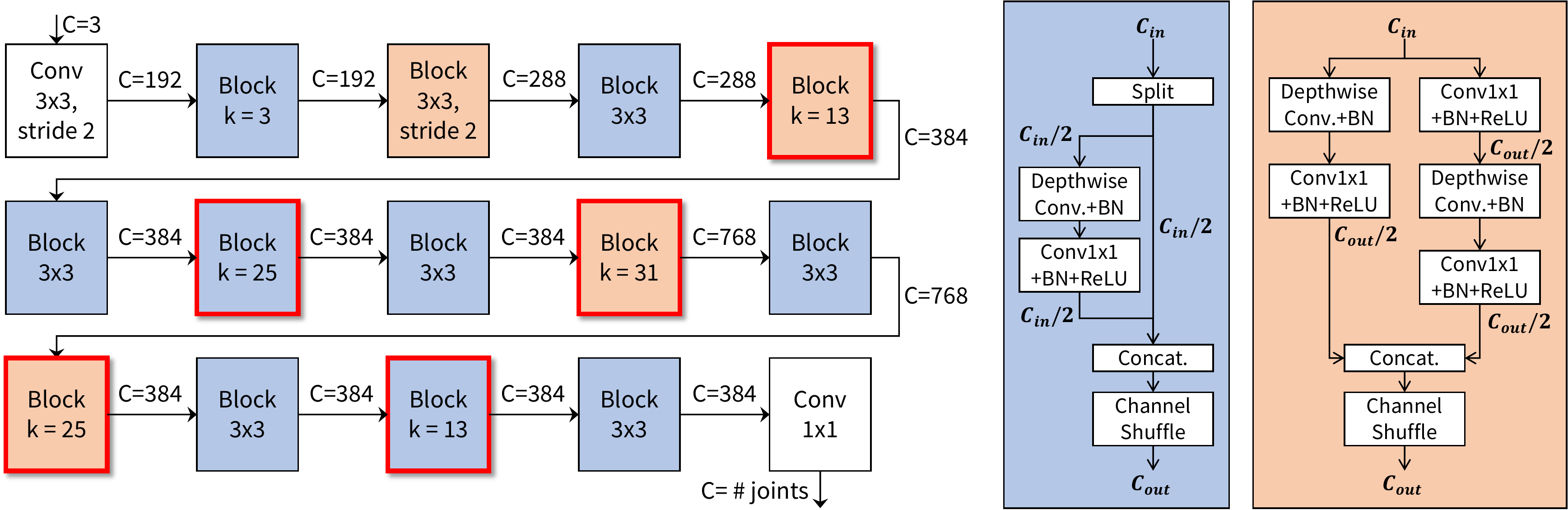}
  \caption{Our dense prediction network. We use blocks with box filters interleaved with blocks with regular 3$\times$3 kernels. The two types of blocks (colored differently) are shown on the right~\cite{ma2018shufflenet}.}
  \label{fig:network_arch}
\end{figure}

\subsection{Dense Prediction Network}
We incorporate box convolution into a network architecture (Figure~\ref{fig:network_arch}) 
with two ``building blocks'' inspired by ShuffleNetV2~\cite{ma2018shufflenet}. 
The blue ``blocks'' only convolve with half of the input channels and ``shuffle''
them with the other, unmodified half.
This diversifies the effective receptive field across channels. 
The orange ``blocks'' are responsible for changing the number of feature 
channels. 
Within the architecture, we interleave regular $3 \times 3$ convolutions with box convolutions 
of various maximum kernel sizes $k$. We employ more channels and larger 
kernels in the middle of the network, reducing the kernel size towards the end 
to produce a sharp prediction. The resolutions of the intermediate feature maps
in our network are the same as the output\,---\,typically a quarter of the input resolution.

\section{Experiments}
We test our method on the task of human pose estimation (more specifically,
estimating the coordinates of each joint).  While it is possible to directly
regress the joint locations~\cite{carreira2016human, toshev2014deeppose},
most recent methods estimate a heatmap for each joint and pick the highest value
as the keypoint~\cite{newell2016stacked,sun2018integral,xiao2018simple,insafutdinov2016deepercut,yang2017learning,sun2019deep}. 
We train and evaluate our method on two datasets: MPII
Human Pose~\cite{andriluka20142d} and Microsoft COCO~\cite{lin2014microsoft}.
To provide supervision for training, the
target heatmap of each joint is generated by centering a Gaussian kernel
with standard deviation equal to 2.0 at the ground-truth location.

\subsection{MPII Human Pose Dataset}

The MPII Human Pose dataset~\cite{andriluka20142d} consists of around 25k images
extracted from online videos. Each image contains one or more people, with
over 40k people annotated in total. Among the 40k samples, $\sim$28k samples are
for training and the remainder are for testing. We use the training/validation
split provided by~\citet{tompson2015efficient}; test annotations are not publicly available.

\header{Evaluation Metrics}
The estimated joint locations are evaluated using the 
Percentage of Correct Keypoints, normalized by head size (PCKh).
Specifically, an estimated joint location
is regarded as correct if its distance from the ground-truth location is no larger
than a constant threshold, normalized by 60$\%$ of the diagonal
of the head bounding box. As is common practice, we report PCKh@0.5.

\header{Training and Testing}
The location and scale of each person in the image is annotated, and we use it
to crop the person and re-size the image to $256 \times 256$, which is a common
practice. We augment the data via random rotation ($\pm30^{\circ}$), scaling
($1.0\pm0.25)$ and flipping.
We train our network using the Adam optimizer~\cite{kingma2014adam}, with the
base learning rate set to $10^{-3}$. The network is trained with a batch size of 64
for 140 epochs. The learning rate is decreased by
a factor of 10 at the 90th and 120th epochs.
Following previous methods~\cite{xiao2018simple,newell2016stacked,chen2018cascaded}, we obtain the
final heatmaps by averaging the outputs for the input image and its flipped version.
The final joint estimation is computed as the location of the highest response,
shifted towards the second-highest response by a quarter of the
distance~\cite{xiao2018simple}.

\header{Results}
Table~\ref{tab:mpii_eval} shows results on the test set, while
Figure~\ref{fig:qualitative_eval}, top row, shows qualitative results. 
SimpleBaseline refers to the publicly available model optimized using a
pretrained ResNet50~\cite{he2016deep} backbone (on the
ImageNet~\cite{deng2009imagenet} classification task).
For a more realistic comparison, we also present results for a version of
SimpleBaseline retrained from scratch. 
Our network, with SAT-accelerated box filters, offers comparable performance
to stacked hourglass networks and SimpleBaseline, while using an order of magnitude
fewer parameters and 1.5-2.5 times less computation.

In addition to comparing against recent well-performing methods, we perform two ablation studies.  First, we replace all of the box filters in the network with conventional
3$\times$3 kernels to evaluate how significantly the receptive field influences
performance.  As shown in Table~\ref{tab:mpii_eval}, second-to-bottom row, the
larger receptive field is critical to good performance, even though each 3$\times$3
kernel has more learnable parameters compared to the box filter (9 vs.\ 4).

We also compare to dilated convolution (third-from-bottom row), utilizing parameters
that match both the computational cost and receptive field of our box filters.
(For example, a box filter of kernel size 13 costs the same and has the same
receptive field as 4$\times$4 convolution with dilation factor 4.)
Despite the significantly greater number of parameters available to dilated convolution (16 vs.\ 4),
our box filters achieve higher performance on every joint, suggesting that SAT-accelerated
box convolution may be an attractive replacement for dilated convolution.
This is likely because each 4$\times$4 dilated convolution kernel essentially 
only utilizes 16 input pixels to produce each output pixel. In contrast, a box 
filter could leverage more input pixels if the learned box is large.

Evaluating the implementation by \citet{burkovbox} on the 
\textbf{test set} is unfortunately infeasible at the time of paper submission.
In the supplemental material, we have included the comparisons on the 
\textbf{validation set}
--- both our method and dilated convolution outperform their implementation.

\begin{table}[t]
  \caption{Comparisons on the MPII Human Pose dataset. 
  ``Pretrain'' indicates that the backbone network is pretrained on the ImageNet classification task.} 
  \centering
  \smallskip
  \newcolumntype{L}[1]{>{\centering\arraybackslash}p{#1}}
  \newcolumntype{C}{>{\centering\arraybackslash}X}
  \setlength\tabcolsep{2.4pt}
  \small
  \begin{tabularx}{\hsize}{Xccccccccccc}
    \toprule
    Method & Pretain& $\#$Params & FLOPs & 
    Head & Shoulder & Elbow & Wrist & Hip & Knee & Ankle & PCKh \\
    \midrule
    8-stage hourglass~\cite{newell2016stacked}& N & 25.1M & 19.1G & 98.2  & 96.3  & 91.2  & 87.1  & 90.1  & 87.4 & 83.6 & 90.9\\
    SimpleBaseline~\cite{xiao2018simple}  & N & 34.0M & 12.0G & 98.0  & 95.3  & 89.1  & 83.9  & 88.3  & 83.7 & 79.1 & 88.7\\
    SimpleBaseline~\cite{xiao2018simple}  & Y & 34.0M & 12.0G & 98.1  & 96.0  & 90.3  & 85.6  & 89.6  & 86.1 & 81.9 & 90.1\\
    Dilated Convolution                       & N & 1.88M & ~~7.7G& 97.7  & 94.6  & 88.8  & 83.5  & 87.4  & 82.3 & 77.7 & 88.0\\
    $3\times3$ Convolution                    & N & 1.87M & ~~7.6G& 95.6  & 92.2  & 82.0  & 75.6  & 76.3  & 69.3 & 63.8 & 80.2\\ \midrule
    Ours                                  & N & 1.85M & ~~7.7G& 98.1  & 95.7  & 90.6  & 85.7  & 89.2  & 84.4 & 79.6 & 89.6\\
    \bottomrule
  \end{tabularx}
  \label{tab:mpii_eval}
\end{table}

\subsection{Microsoft COCO Keypoint Detection}

\begin{table}[t]
  \newcolumntype{L}[1]{>{\centering\arraybackslash}p{#1}}
  \newcolumntype{C}{>{\centering\arraybackslash}X}
  \caption{
      Comparisons on the \texttt{val2017} set of MS COCO. 
      ``Pretrain" --- backbone is pretrained on 
      ImageNet~\cite{deng2009imagenet}; 
      ``OHKM'' ---  Online Hard Keypoints Mining;
      ``-'' ---  data is not publicly available.
  }
  \label{tab:coco_val}
  \centering
  \smallskip
  \setlength\tabcolsep{2pt}
  \footnotesize
  \begin{tabularx}{\hsize}{Xp{13mm}cccccccccc}
    \toprule
    Method & Backbone & Input size & Pretrain & $\#$Params & FLOPs & AP & AP$^{50}$ & AP$^{75}$ & AP$^{M}$ & AP$^{L}$ & AR \\   
    \midrule
    8-stage hourglass~\cite{newell2016stacked} & - & 256$\times$192 
    & N & 25.1M & 14.3G&
    66.9& -& -& -& -& -\\
    CPN~\cite{chen2018cascaded} & ResNet50 & 256$\times$192
    & Y & 27.0M & ~~6.2G&
    68.6& -& -& -& -& -\\
    CPN + OHKM~\cite{chen2018cascaded} & ResNet50& 256$\times$192
    & Y & 27.0M & ~~6.2G&
    69.4& -& -& -& -& -\\
    SimpleBaseline~\cite{xiao2018simple} & ResNet50& 256$\times$192
    & N & 34.0M & ~~8.9G&
    69.3& 88.3& 77.0& 66.2& 75.8& 75.3\\
    SimpleBaseline~\cite{xiao2018simple} & ResNet50& 256$\times$192
    & Y & 34.0M & ~~8.9G&
    70.4& 88.6& 78.3& 67.1& 77.2& 76.3\\
    SimpleBaseline~\cite{xiao2018simple} & ResNet152& 256$\times$192
    & Y & 68.6M & 15.7G&
    72.0& 89.3& 79.8& 68.7& 78.9& 77.8\\
    \midrule
    Ours & - &  256$\times$192
    & N & 1.85M & ~~5.8G&
    69.9& 88.6& 76.7& 66.1& 76.5& 75.1\\
    \bottomrule
  \end{tabularx}
  \caption{
      Comparisons on the \texttt{test-dev2017} set of Microsoft COCO.
  }
  \label{tab:coco_test_dev}
  \smallskip
  \setlength\tabcolsep{2pt}
  \begin{tabularx}{\hsize}{Xp{23mm}cccccccccc}
    \toprule
    Method & Backbone & Input size & Pretrain & $\#$Params 
    & AP & AP$^{50}$ & AP$^{75}$ & AP$^{M}$ & AP$^{L}$ & AR \\
    \midrule
    OpenPose~\cite{Cao_2017_CVPR} & - & -
    & - & - &
    61.8& 84.9& 67.5& 57.1& 68.2& 66.5\\
    Mask-RCNN~\cite{he2017mask} & ResNet-50-FPN & -
    & - & - &
    63.1& 87.3& 68.7& 57.8& 71.4& - \\
    Integral Pose~\cite{sun2018integral} & ResNet101 & 256$\times$256
    & - & 45.0M & 
    67.8& 88.2& 74.8& 63.9& 74.0& - \\
    SimpleBaseline~\cite{xiao2018simple} & ResNet50 & 256$\times$192
    & N & 34.0M &
    68.8& 90.3& 76.8& 65.7& 74.5& 74.5\\
    SimpleBaseline~\cite{xiao2018simple} & ResNet50 & 256$\times$192
    & Y & 34.0M &
    70.0& 90.9& 77.9& 66.8& 75.8& 75.6\\
    CPN~\cite{chen2018cascaded} (ensembled) & ResNet-Inception & 384$\times$288
    & - & - &
    73.0& 91.7& 80.9& 69.5& 78.1& 79.0\\
    SimpleBaseline~\cite{xiao2018simple} & ResNet152 & 384$\times$288
    & Y & 68.6M &
    73.7& 91.9& 81.1& 70.3& 80.0& 79.0\\
    \midrule
    Ours & - & 256$\times$192
    & N & 1.85M &
    68.9& 90.3& 76.0& 65.3& 74.9& 74.3\\
    \bottomrule
  \end{tabularx}  
\end{table}

Multi-person pose estimation is one of the tasks on 
the Microsoft COCO dataset~\cite{lin2014microsoft}, which contains over 200k images and 250k person instances.
Each person is annotated with 17 joints. 
For training, we use the \texttt{train2017} split, which has $\sim$57k images and 
$\sim$150k samples.
Validation and evaluation are performed on the \texttt{val2017}
($\sim$5k images) and the \texttt{test-dev2017} ($\sim$20k images) splits,
respectively.

\header{Evaluation Metrics}
The performance is evaluated using Object Keypoint Similarity (OKS), which is
defined as $\nicefrac{\sum_{i} \exp \left(-d_{i}^{2} / 2 s^{2}
\kappa_{i}^{2}\right) \delta\left(v_{i}>0\right)}{\sum_{i}
\delta\left(v_{i}>0\right)}$, where $d_i$ is the Euclidean distance from the
i-th predicted joint location to ground truth, $v_i$ is the visibility flag,
and $\kappa_i$ is a per-keypoint constant that controls falloff. We report the
average precision (AP) and recall (AR). Average precision is evaluated for OKS =
0.5 (AP$^{0.5}$), OKS = 0.75 (AP$^{0.75}$), OKS = 0.5:0.05:0.95 (AP), medium
objects (AP$^M$) and large objects (AP$^L$). Average recall is evaluated for OKS
= 0.5:0.05:0.95.

\header{Training}
We follow similar training and testing procedures as above,
but with slight differences. The input is cropped and
re-sized to 256$\times$192. Because the dataset aims for multi-person pose
estimation, and the location of each person is not available during testing, we
adopt the Faster-RCNN human detector~\cite{girshick2015fast},
which is also used by SimpleBaseline~\cite{xiao2018simple}. During training, we
still rely on the ground-truth bounding boxes to determine the location of each
sample. The network is trained for 210
epochs, and we decrease the learning rate by a factor of 10 at the 170th and
200th epochs.

\header{Results}
Table~\ref{tab:coco_val} shows the results on the \texttt{val2017} set.
Our method outperforms the SimpleBaseline (ResNet50 backbone) trained from scratch,
and even performs slightly better than CPN, which uses both online hard keypoint mining
and a pretrained backbone. 
Note that the number of parameters in our network is an order
of magnitude lower than previous methods. Table~\ref{tab:coco_test_dev}
demonstrates that we also achieve competitive results on the
\texttt{test-dev2017} set, when compared to existing methods.
Figure~\ref{fig:qualitative_eval}, bottom row, shows qualitative results on 
the \texttt{test-dev2017} set.

\subsection{Analysis}
\begin{figure}[t]
    \centering
    \newcolumntype{C}{>{\centering\arraybackslash}X}
    \setlength\tabcolsep{5pt}
    \begin{tabularx}{\hsize}{CCCCC}
        \includegraphics[width=\hsize]{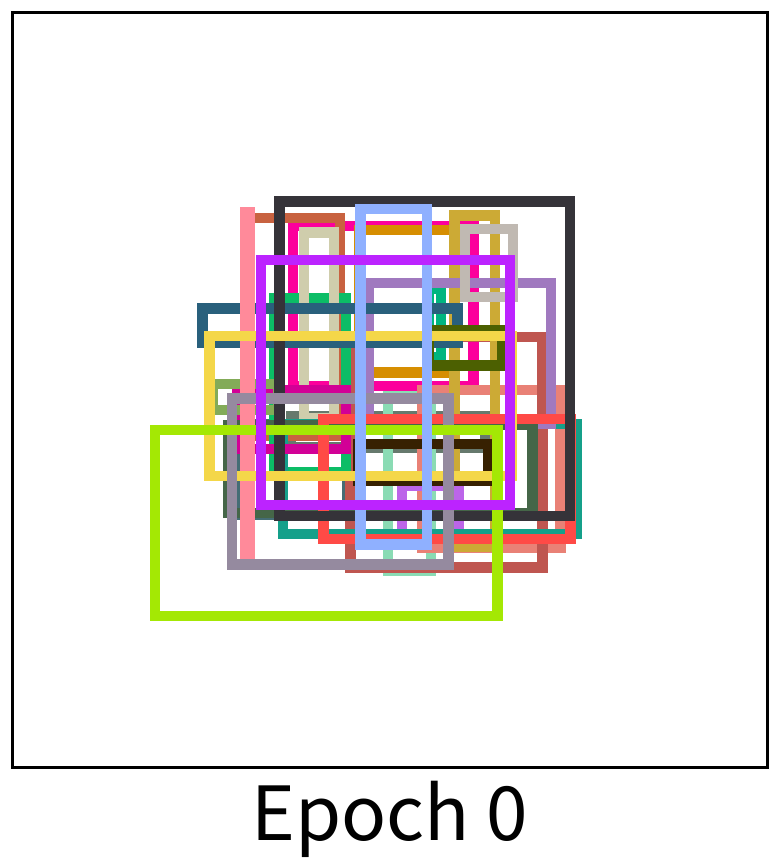}&
        \includegraphics[width=\hsize]{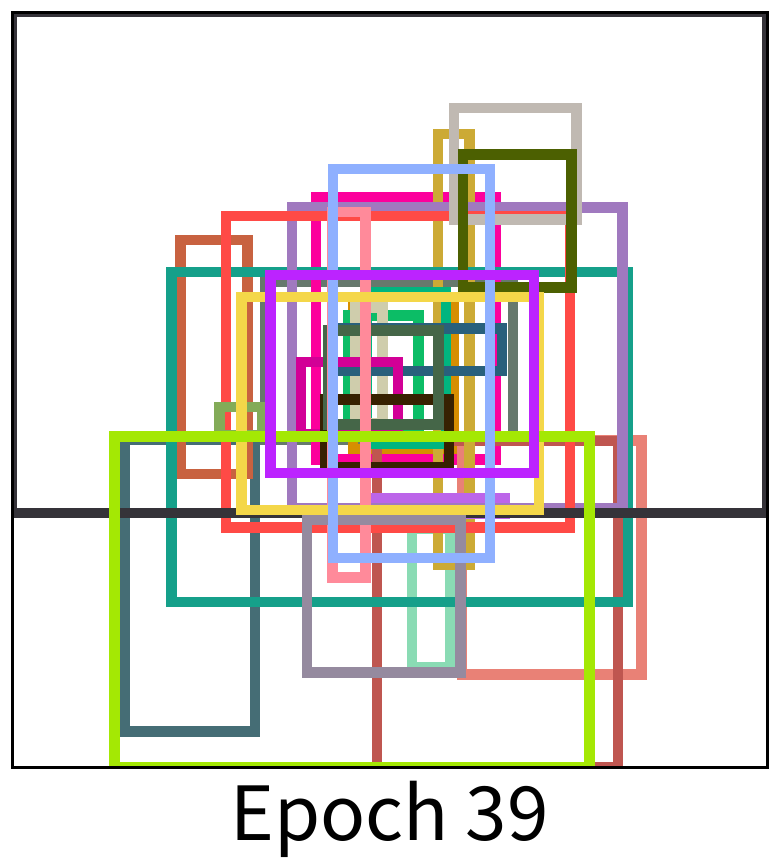}&
        \includegraphics[width=\hsize]{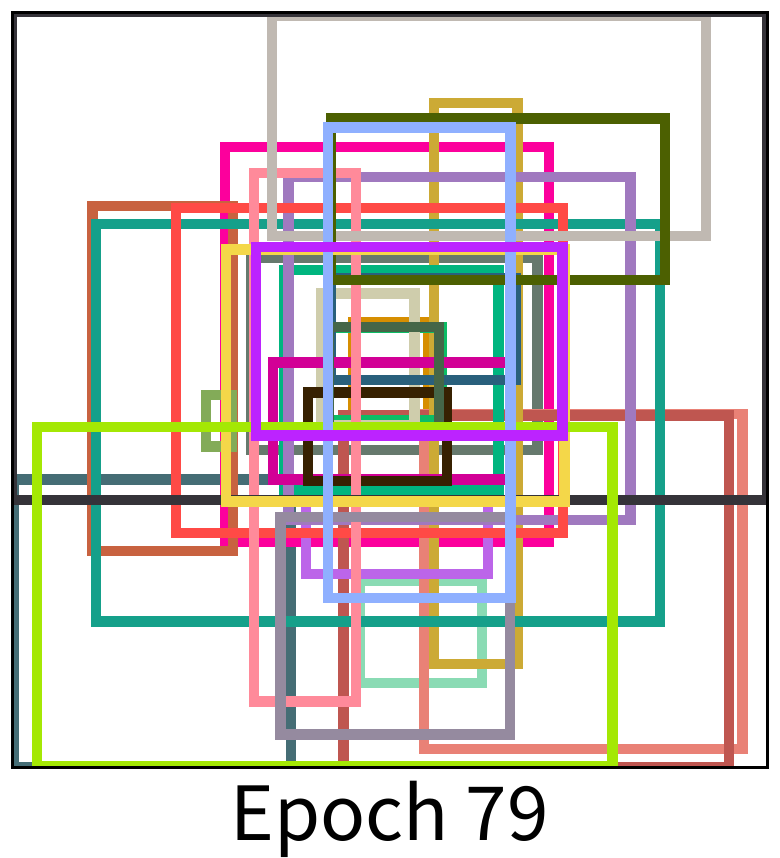}&
        \includegraphics[width=\hsize]{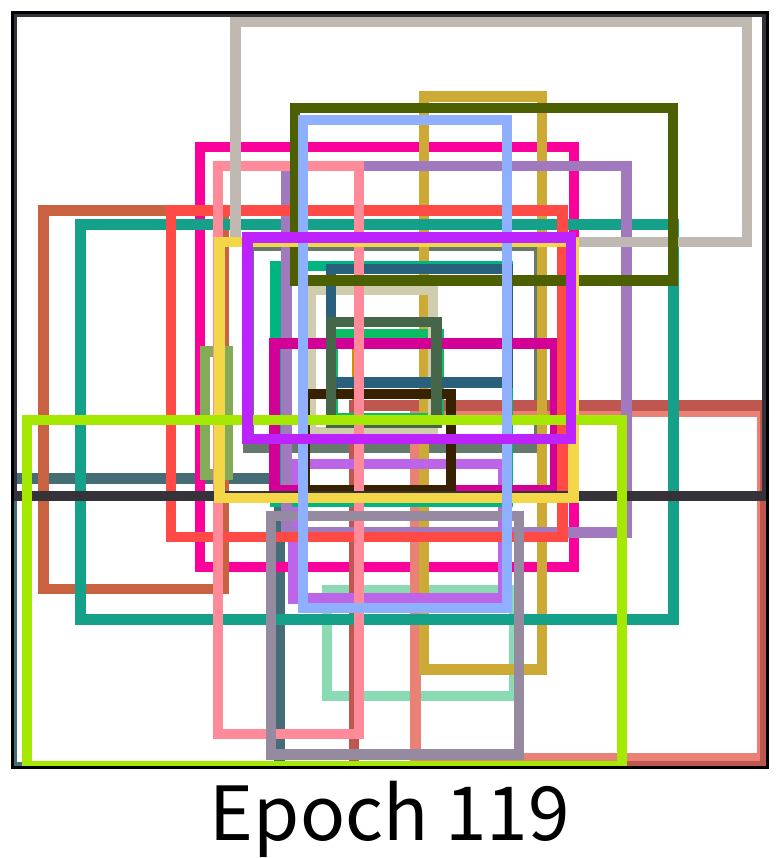}&
        \includegraphics[width=\hsize]{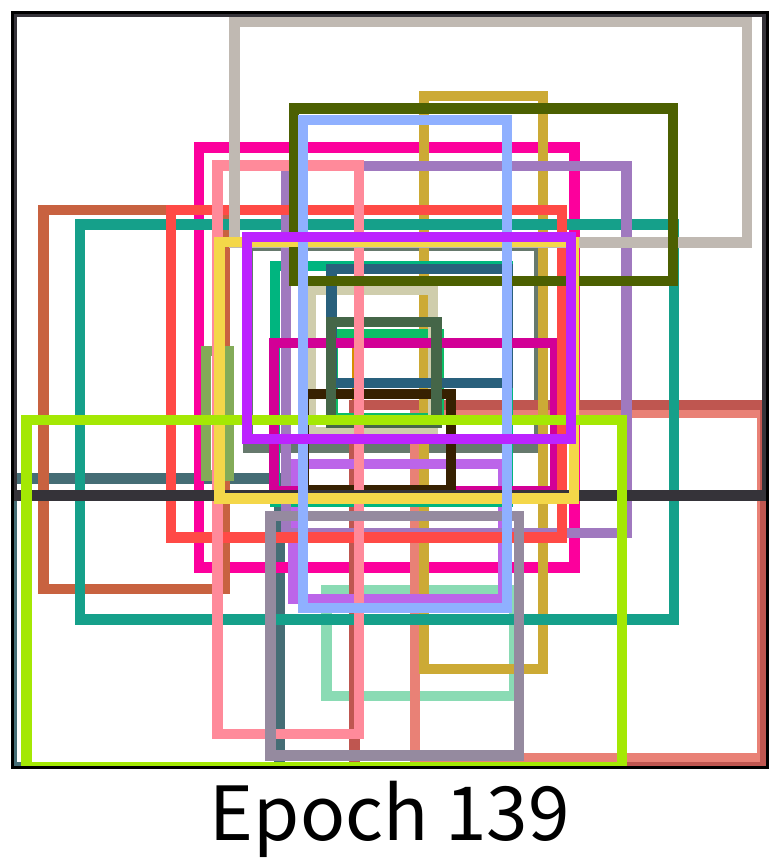}
    \end{tabularx}
    \caption{Evolution of learned boxes. As training proceeds, the boxes become more diverse, producing varied features maps. The learning rate is reduced at the 90th and 120th epochs.}
    \label{fig:boxes_visualization}
\end{figure}
\header{Learned Filters}
Figure~\ref{fig:boxes_visualization} shows a subset of boxes (32 instances)
learned in the last
layer that uses box filters ($13 \times 13$ kernel) at different stages of 
training. \citet{burkovbox} observe
that a certain number of boxes learned in their network shrink to the minimal
size under the imposed L2-regularization. Such regularization is not necessary in
our approach since the size of each box is bounded by a pre-defined kernel. We
therefore do not observe any similar effect. They also demonstrate a
counter-intuitive phenomenon that their learned boxes tend to be \emph{symmetric
w.r.t.\ the vertical axis}, and ``even when horizontal flip augmentations are
switched off during training''. 
The learned boxes after convergence in Figure~\ref{fig:boxes_visualization}
show that we do not encounter this issue. We therefore conclude that four
parameters are still necessary to represent each box.

\begin{figure}[t]
    \centering
    \newcolumntype{C}{>{\centering\arraybackslash}X}
    \setlength\tabcolsep{10pt}
    \begin{tabularx}{0.75\hsize}{CC}
        \includegraphics[width=\hsize]{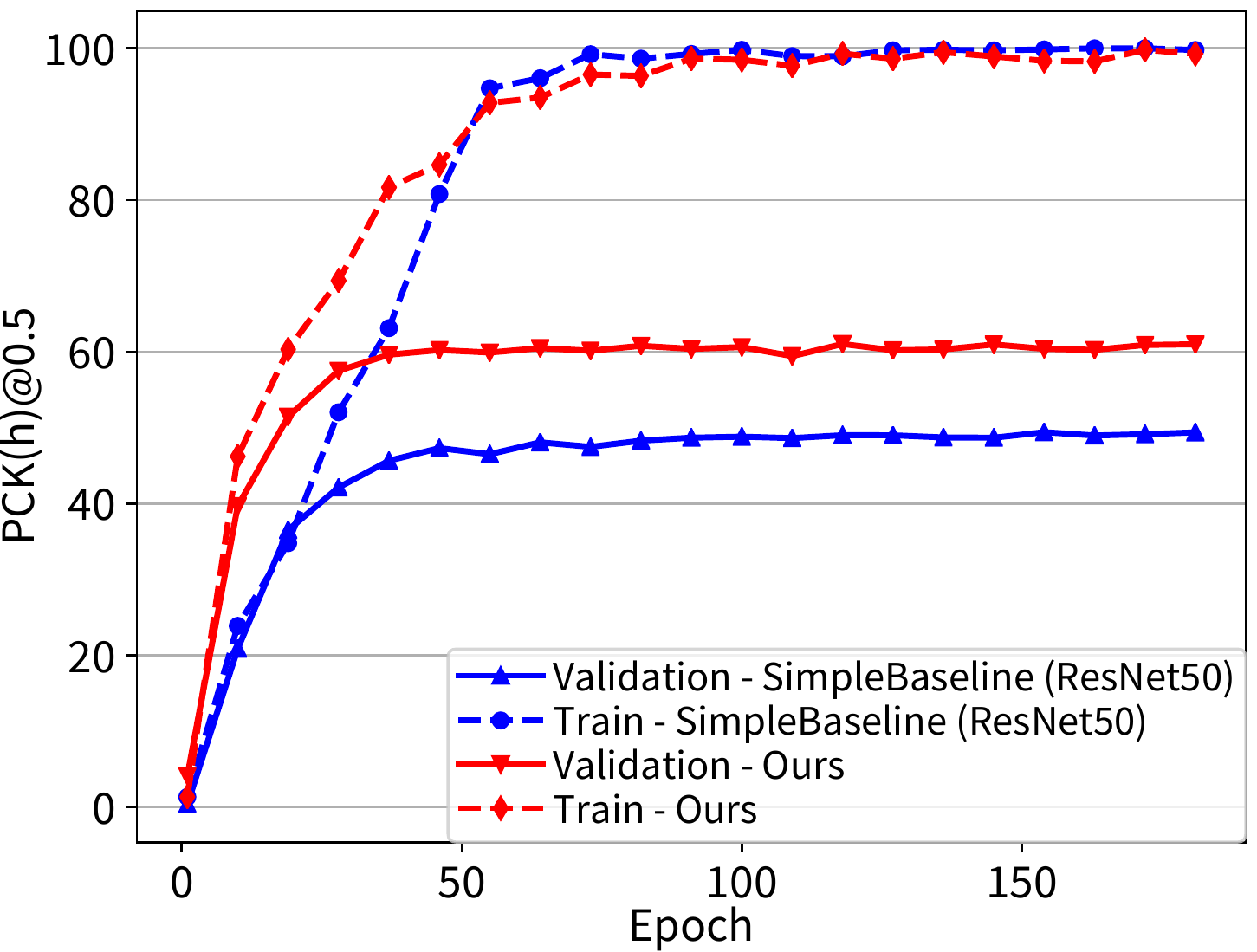}&
        \includegraphics[width=\hsize]{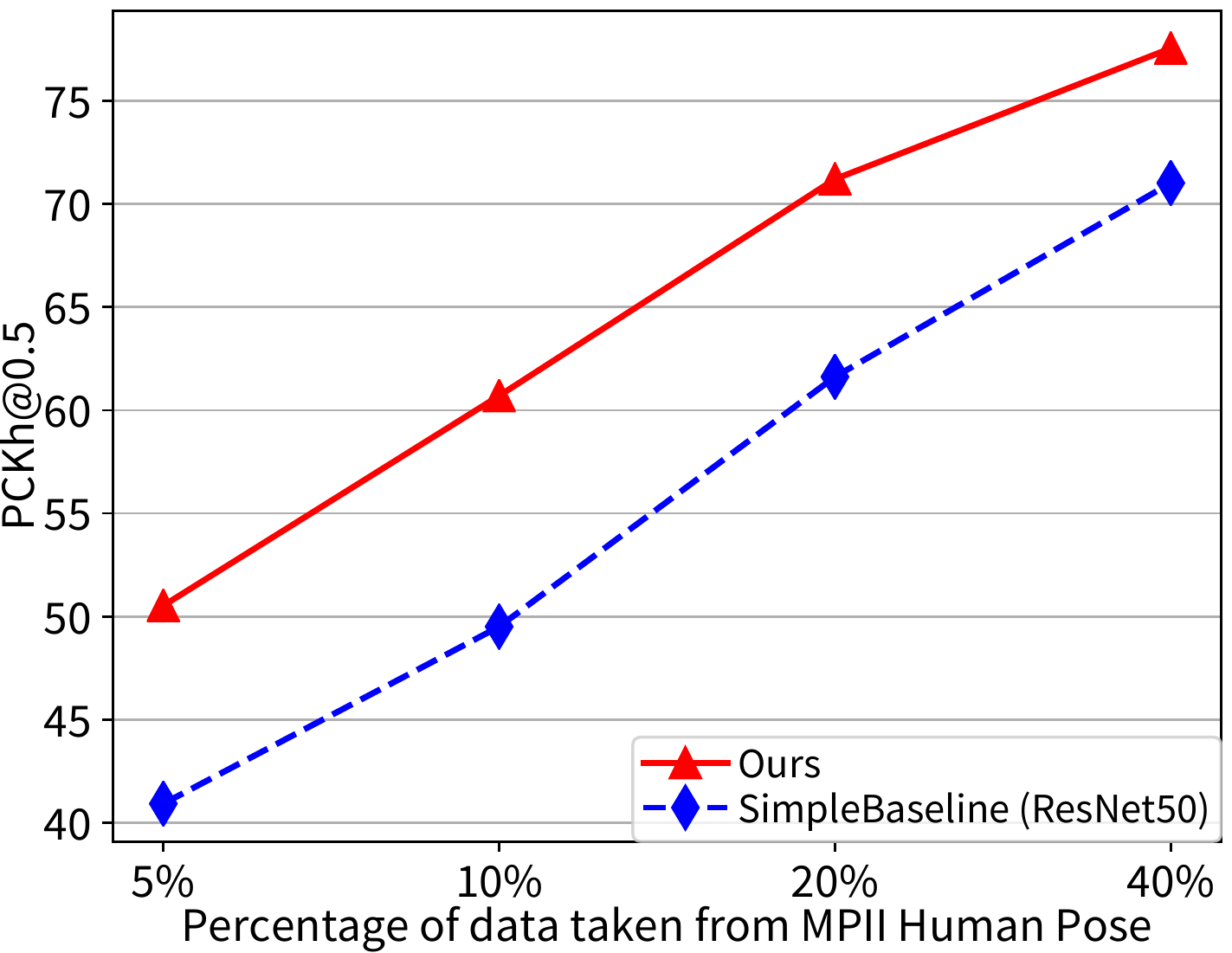}
    \end{tabularx}
    \caption{Overfitting analysis on the MPII Human Pose dataset. 
    \textbf{Left:} training and validation accuracy of our method and the SimpleBaseline. Training accuracy refers to PCK@0.5 (without head-size normalization since each training sample has been re-sized to 256$\times$256 already).
    \textbf{Right:} validation accuracy after training saturates, using different amounts of training data.}
    \label{fig:overfitting}
\end{figure}
\header{Overfitting Analysis}
Complex neural networks are known to be more likely to overfit to small
datasets. A kernel used to convolve with a small feature map would
be \emph{poorly utilized}, unless the size of the training data is large. While
our network has many fewer parameters than existing architectures, the
utilization of each kernel is high since we maintain high-resolution feature
maps. This implies that our method may generalize better to new
data when the training data is insufficient. To verify this, we take a small
subset (10$\%$) of the MPII Human Pose dataset and allow the networks to overfit
to it. To make overfitting happen more quickly, we disable random
rotation/scaling and keep flip augmentation only. Figure~\ref{fig:overfitting},
left, shows the training progress of our network and the SimpleBaseline (using
a ResNet50 backbone). Surprisingly, the gap in validation accuracy between
SimpleBaseline and our method is significant after both networks saturate (i.e.,
achieving $\approx$100$\%$ training accuracy). While our network converges
faster at the beginning, it takes longer to \emph{completely} overfit to the
training data, which may explain the higher validation accuracy after convergence.
We have tried varying the learning rate for training the SimpleBaseline, but
the eventual validation accuracy is nearly unaffected.
In Figure~\ref{fig:overfitting}, right, we show comparisons using different
amounts of training data. Our method consistently outperforms the SimpleBaseline.

\section{Conclusions and Discussion}
In this work, we propose a large-kernel convolution layer leveraging summed-area
tables to accelerate computation. Using the proposed layer, we design an
end-to-end differentiable dense prediction network that produces pixel-level
prediction while ensuring a large effective receptive field. We demonstrate
through the human pose estimation task that our method leads to competitive
performance using many fewer parameters and at lower computational cost.

Our network maintains high-resolution intermediate feature maps only, which
consume more GPU memory than previous networks using downsampling. As
mentioned previously, reducing the number of channels would hurt the diversity
of box filters, thus making performance worse. A solution to this is the kernel
splitting strategy we have mentioned in Section~\ref{sec:SAT_conv}, which trades
more \texttt{multadds} for a lower memory footprint. 
Nevertheless, once the learned boxes converge, we can round the sampling points
(i.e., box corners) to
integers and only fine-tune the rest of the network. Sampling using integer
coordinates does not require interpolation, which would use many fewer \texttt{multadd} 
operations.

{
\small
\bibliographystyle{plainnat}
\bibliography{neurips}
}
\end{document}